\documentclass[conference]{IEEEtran}
\IEEEoverridecommandlockouts
\usepackage{graphicx} 
\usepackage{balance}
\usepackage{multirow}
\usepackage{array}
\usepackage{subfig}
\usepackage{hyperref}
\usepackage{amsmath,amssymb,amsfonts}

\usepackage{algorithmic}
\usepackage{textcomp}
\usepackage{xcolor}
\def\BibTeX{{\rm B\kern-.05em{\sc i\kern-.025em b}\kern-.08em
    T\kern-.1667em\lower.7ex\hbox{E}\kern-.125emX}}

\begin{document}

\title{Towards bio-inspired unsupervised representation learning for indoor aerial navigation}

\author{
    \IEEEauthorblockN{Ni Wang\IEEEauthorrefmark{1}, Ozan \c{C}atal\IEEEauthorrefmark{1}, Tim Verbelen\IEEEauthorrefmark{1}, Matthias Hartmann\IEEEauthorrefmark{2}, Bart Dhoedt\IEEEauthorrefmark{1}}
    \IEEEauthorblockA{\IEEEauthorrefmark{1}IDLab, Ghent University - imec, Belgium
    \\firstname.lastname@ugent.be}
    \IEEEauthorblockA{\IEEEauthorrefmark{2}imec, Belgium
    \\firstname.lastname@imec.be}
}

\maketitle

\begin{abstract}
Aerial navigation in GPS-denied, indoor environments, is still an open challenge. Drones can perceive the environment from a richer set of viewpoints, while having more stringent compute and energy constraints than other autonomous platforms. To tackle that problem, this research displays a biologically inspired deep-learning algorithm for simultaneous localization and mapping (SLAM) and its application in a drone navigation system. We propose an unsupervised representation learning method that yields low-dimensional latent state descriptors, that mitigates the sensitivity to perceptual aliasing, and works on power-efficient, embedded hardware. The designed algorithm is evaluated on a dataset collected in an indoor warehouse environment, and initial results show the feasibility for robust indoor aerial navigation.
\end{abstract}


\section{Introduction}
One of the major concerns of autonomous drone navigation is the ability to safely fly in an unknown space, and keep track of the drone's location. Traditional simultaneous localization and mapping (SLAM) approaches tackle this problem by building a (dense) metric map of the environment, and then tracking your pose within that map~\cite{Labbe2019}. Although successful for mobile robots, the problem for drones is more complex due to the extra degrees of freedom in possible viewpoints, and due to the limited power and compute budget available. Also, estimated odometry on a drone platform is typically more noisy than odometry estimated from wheel encoders from a mobile robot.

A different take on SLAM is a more bio-inspired approach that is based on recent findings on the inner workings of the hippocampal brain area in mammals. In this case, poses are represented as activations of grid and head direction cells in a continuous attractor network, and linked to representations of visual percepts~\cite{rat}. The resulting map now becomes a graph with pose and visual representations, which is no longer a full metric representation, but rather a topological map. 

In recent work~\cite{latent} we presented LatentSLAM, which further improved upon these methods by using a learnt latent representation of visual percepts, adopting state of the art deep learning models. We train these latent representations in an unsupervised manner, by predicting future observations, in similar vein how brains supposedly update themselves through predictive coding~\cite{PMID:10195184}. We illustrated how this yields more robust place descriptors that enable to mitigate perceptual aliasing, by evaluating on a mobile robot in a warehouse environment~\cite{latent}.

In this paper, we further extend LatentSLAM to work for drone navigation, in particular coping with a noisy odometry signal, and assuring that our latent space model can be evaluated in real-time on an embedded compute platform. We present some initial results on a dataset collected with a real drone platform in a challenging indoor environment.

\begin{figure*}[]
 \centering
 \includegraphics[width=0.95\textwidth]{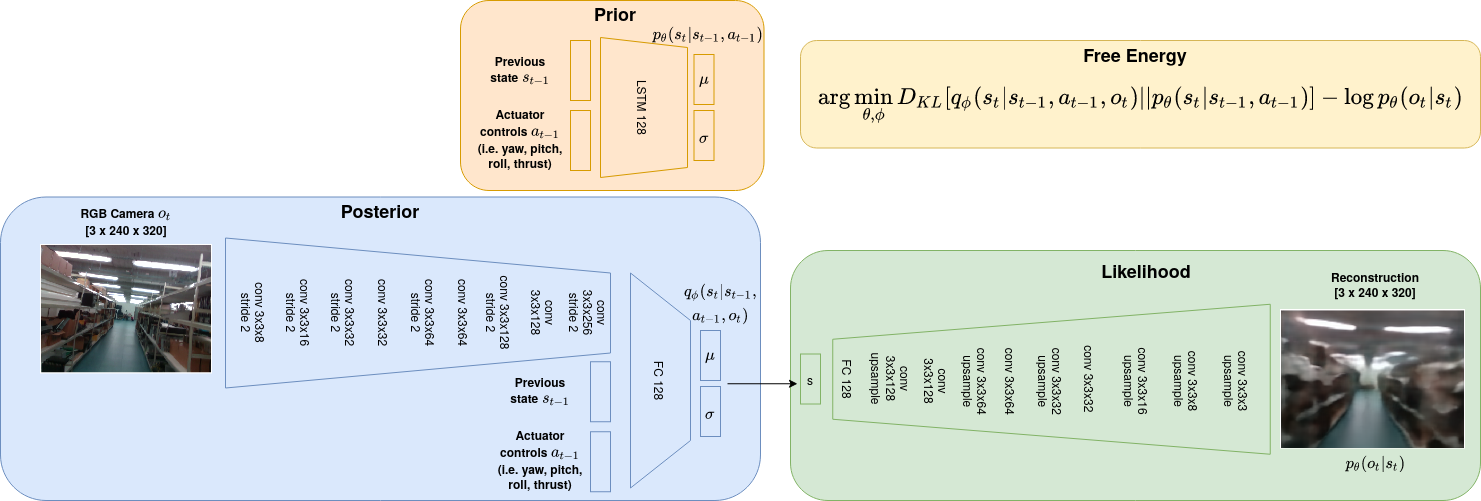}
 \caption{Our generative model is composed out of three blocks. The Prior block needs to learn latent space dynamics, predicting the current state from previous state and actuator controls. The Posterior block combines the previous state and controls with features from a convolutional pipeline to make a posterior estimate of the current state. Finally, the likelihood model reconstructs the observation from state samples. These blocks are trained end-to-end by minimizing free energy, and at inference time only the Posterior network is used for inferring latent states.}
 \label{fig:architecture}
\end{figure*}

\section{Related Work}
Indoor aerial navigation demands real-time availability and processing of perceptual information for understanding the surrounding environment. 
For the past decades, a lot of research works have shown great efforts in SLAM for aerial agents, combining both geometric and semantic information for drone navigation. 
For example, in \cite{imu} a localization method was proposed by integrating monocular SLAM and inertial measurement unit (IMU) data. Similarly, another work fused IMU data and a `ORB-SLAM' for aerial navigation in GPS-denied environment, where `ORB-SLAM' uses Oriented Fast and rotated BRIEF (ORB) features \cite{orb}.
`Air-SSLAM', a visual SLAM method which exploits a stereo camera configuration, starts from computing keypoints and the correspondent descriptors over the pair of images, using good features-to-track and rotated-binary robust-independent elementary features, respectively, and afterwards creates the map \cite{air}.
Recently, `VPS-SLAM' proposes a lightweight and real-time visual semantic SLAM framework adaptive to aerial robotic platforms by combining low-level visual/visual-inertial odometry (VO/VIO) along with geometrical information corresponding to planar surfaces extracted from detected semantic objects \cite{vps}. 
Another group of researchers built an extended drone visual/inertial SLAM approach by utilizing internal landmarks that are created by the algorithm itself, in addition to opportunity-based features that are used by standard SLAM solutions \cite{drone}. 

Although using basic visual features for visual SLAM is beneficial from a low power processing perspective, these approaches typically fail to disambiguate similar looking percepts. This is typically the case in indoor environments such as warehouses, where the same features appear over and over again in different aisles. One approach is to build a dense metric map of features, typically a dense point cloud, but this comes at the cost of extra compute and memory. An other method is to build a topological map of the environment, as proposed by `RatSLAM', a computational model based on the neural encoding of space by the rodent hippocampus, which enabled real-time SLAM on a two-wheeled robot `Pioneer' \cite{rat}. Based on this, we recently proposed `LatentSLAM', an extension of RatSLAM which uses a learnt, compact latent representation to disambiguate different locations \cite{latent}. In this work, we further investigate to what extent this approach can also be applied to drone navigation.

\section{Latent SLAM}

LatentSLAM is an extension of RatSLAM, a bio-inspired visual SLAM algorithm based on the rodent hippocampus. It consists of three main component: view cells, pose cells and the experience map. View cells are linked to a visual template, and become active when the current camera input matches the template. If a view cell becomes active, it activates a corresponding pose cell, that represents the pose at which the template was initially created. These pose cells are aligned in a 3D cube, and form a continuous attractor network (CAN) that track the robots pose in terms of its x- and y-coordinates, as well as orientation. Pose cells are activated either by odometry updates that shift activity, or by view cells that inject activity when matched. This allows the pose CAN to combine odometry and visual information, resulting into a pose estimate reflected by the highest active pose cell.
Finally, view cells and their corresponding pose are structured as nodes in an experience map. When a new camera frame is processed, and it is not matched within a certain threshold with an existing view cell, a new view cell is generated, and a new node is added to the experience map. Whenever a camera frame is matched, a link is added in the experience map linking the previously active node with the newly matched node. This way, the experience map forms a topological view of the environment and loop closures can be detected. For a more in depth treatment of RatSLAM, we refer to~\cite{rat}.

In LatentSLAM, we extend the RatSLAM algorithm by not using visual template matching for the view cells. Instead, we learn a latent state representation using deep neural networks for mapping camera frames into a low-dimensional state space. An overview of the model is given in Figure~\ref{fig:architecture}. Given a set of observations (i.e. camera images) and actions (i.e. drone thrust commands), we want to learn a latent representation $s$ that captures all important information. To do so, we define three models: a prior network, a posterior network and a likelihood network. The posterior network infers the current state given the previous state, action and latest observation. The likelihood model on the other hand, reconstructs the observation from state samples. The posterior distribution is regularized by a prior network, which predicts the current state from the previous state and action only. This ensures that our latent state space captures both the correct dynamics, as well as the visual features from the observation. We train these networks end-to-end by minimizing the so-called free energy \cite{Friston2016}, or equivalently maximizing the evidence lower bound~\cite{kingma,Rezende2014}. 

\begin{equation}
\begin{split}
\mathfrak{L} &= \sum_{t} D_{KL} [q_{\phi} (s_{t} |s_{t-1}, a_{t-1}, o_{t}) \mid \mid p_{\theta}(s_{t} | s_{t-1}, a_{a-1})] \\ & - \log p_{\theta}(o_t |s_t)
\end{split}
\end{equation}

\begin{figure*}[t]
\centering
 \subfloat[][Test trajectory]{
   \includegraphics[height=45mm]{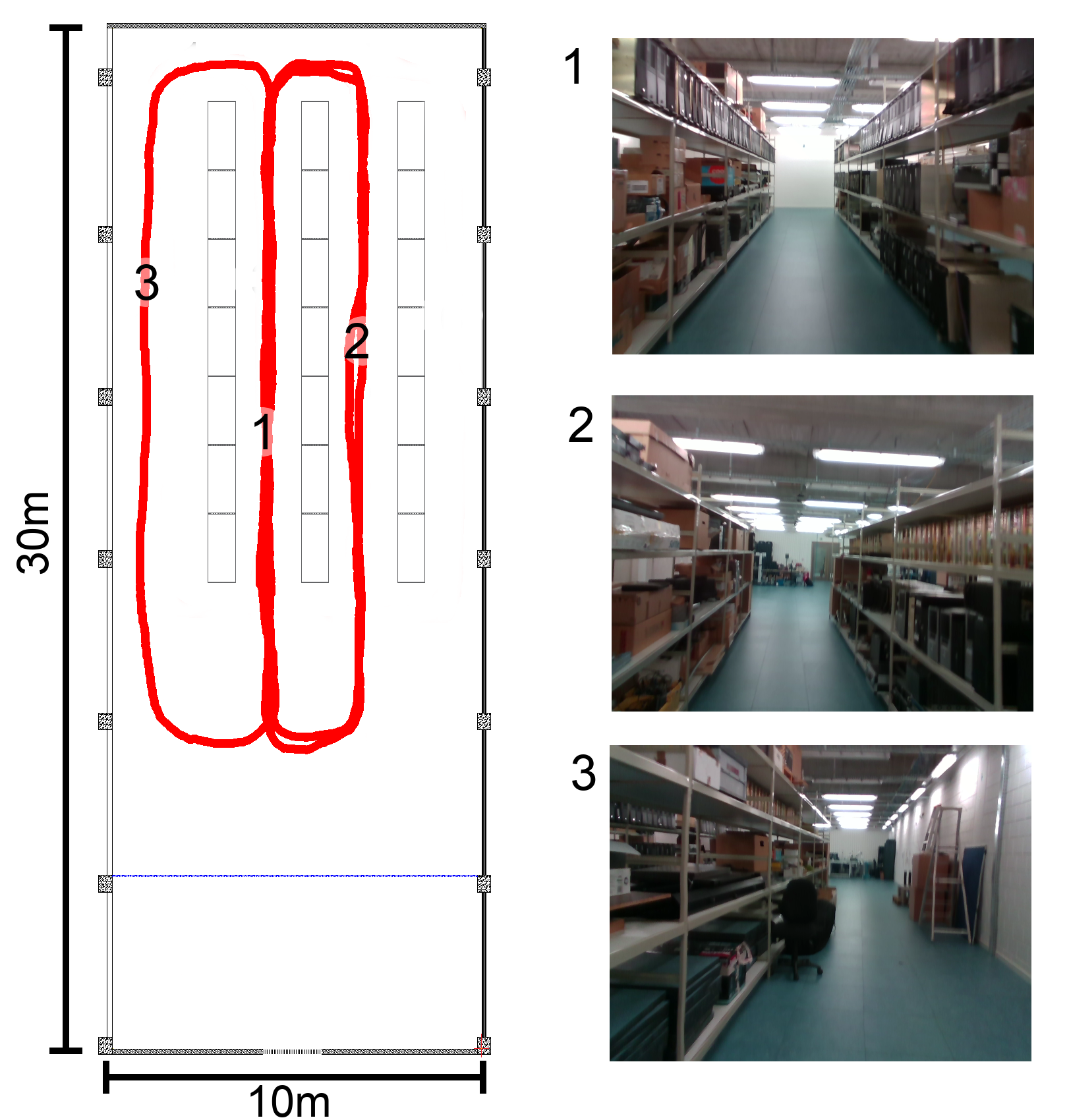}
   \label{fig:track}
 }
 \subfloat[][Odometry]{
   \includegraphics[height=45mm]{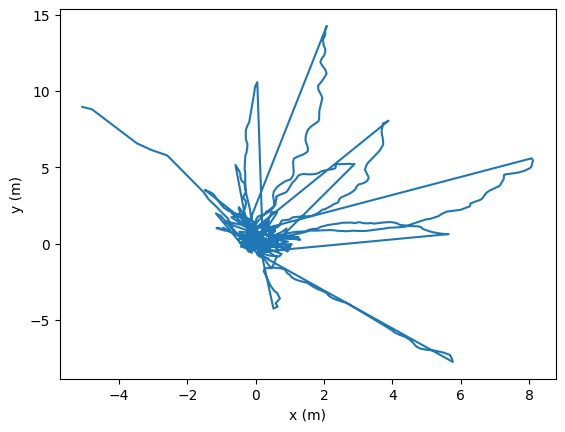}
   \label{fig:odometry}
 }
 \subfloat[][LatentSLAM map]{
   \includegraphics[height=45mm]{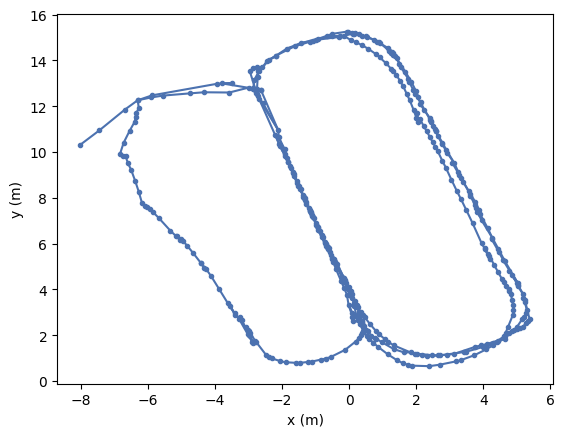}
   \label{fig:map}
 }
 \caption{We fly our drone in a warehouse-like lab environment, traversing similar looking aisles with racks and shelves (a). The raw odometry signal is very noisy, and resets to origin when the optical flow tracker loses tracking (b). However, our LatentSLAM algorithm is still able to recover a correct topological map of the environment.}
\end{figure*}

At inference time, only the posterior model is used to infer the belief state distribution for each observation, and the mean of this distribution is to serve as template for a local view cell. For matching view cells, we now use the cosine distance between the latent code of the current camera frame and the latent code corresponding to a view cell, and match if this falls below a certain threshold. More details on our approach can be found in~\cite{latent}.

\section{Results}

\begin{figure}[b!]
    \centering
    \includegraphics[width=0.4\textwidth]{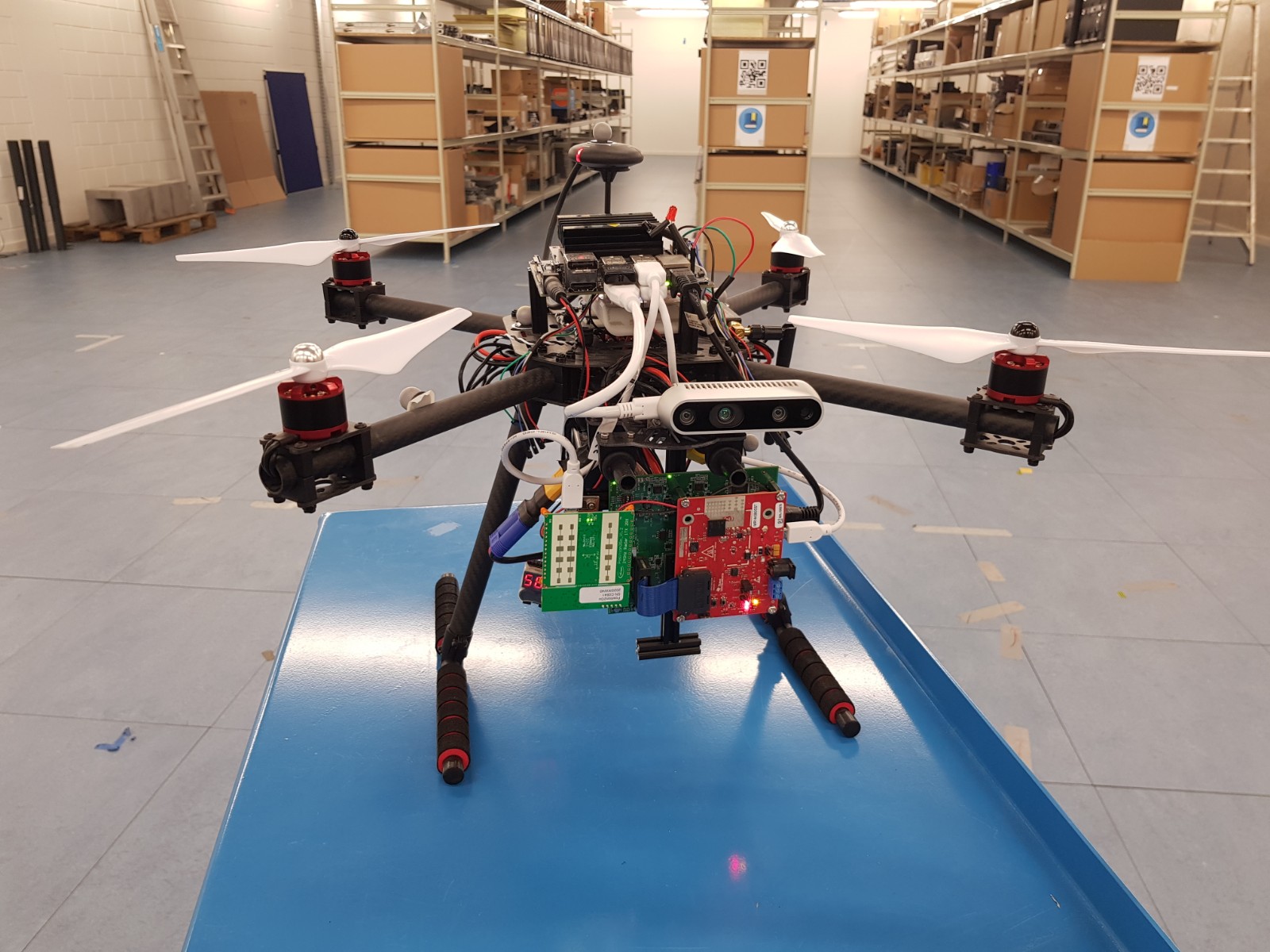}
    \caption{Our experimental NXP drone platform, equipped with RGBD camera, a 24GHz Infineon radar, a 79GHz TI radar and a Jetson nano compute platform. In this paper, we focus on RGB camera data.}
    \label{fig:drone}
\end{figure}

To evaluate our LatentSLAM approach for indoor drone navigation, we conducted experiments in our industrial IoT lab environment, a 300 $\text{m}^2$ environment consisting of racks and shelves mimicking a warehouse. This is a challenging environment, as different aisles look very similar, making them hard to disambiguate from perception. Our drone is an NXP Hovergames platform with a px4 flight-controller, mounted with an RGBD camera and radars as shown on Figure~\ref{fig:drone}. In this paper, we only use the RGB frames, but as future work we will consider the extra sensor modalities. We also equipped the drone with a Jetson Nano module as additional compute resource.

We first collected data from 7 flights, totalling about 7500 frames, on which we train our model as described in Section~3. We train for 500 epochs using the Adam optimizer with initial learning rate 1e-4. After training, we use the Posterior model for estimating latent states from observations at inference time. We evaluate on a separate test flight, consisting of 1000 frames flying loops through three of the aisles as depicted in Figure~\ref{fig:track}. As odometry source, we use the Extended Kalman Filter state from the drone flight controller, which integrates information from IMU, gyroscopes, compass and a px4flow optical flow sensor. It is clear from Figure~\ref{fig:odometry} that this signal is very noisy, and the position estimate often resets to the origin whenever the optical flow tracker loses tracking. Despite this, our LatentSLAM system is still able to recover a correct topological map of the environment, correctly matching observations and loop closures as shown in Figure~\ref{fig:map}.

To enable real-time execution on a battery-powered drone platform, inference of our posterior model only takes 25ms on a Jetson Nano board. Our latent space currently has 32 dimensions, which results in a low memory footprint for building and maintaining the map.

Note that we still use a 3-DOF `    pose representation, which seems rather limited for a drone. However, we argue that for indoor environments explicitly modeling the height, pitch and roll does not add much information for correct mapping and loop closure. Although we did not fly at a fixed height during the test sequence, our system correctly matched corresponding locations. It might be interesting to investigate whether it modeled these extra degrees of freedom in the latent space.  

\section{Conclusions \& Future Work}
In this paper, we extended LatentSLAM, a bio-inspired SLAM method using unsupervised representation learning for aerial navigation in indoor environment. We demonstrated satisfactory performance in mapping and navigation on a real-world drone platform in presence of noisy odometry signal, while being able to execute on an embedded computation platform.
As future work we plan to further extend our dataset and benchmark against state-of-art of existent SLAM methods. We also plan to use the additional depth and radar data to further increase the robustness, i.e. in low light or smoke conditions. 

\section*{Acknowledgment}

O.\c{C}. is funded by a Ph.D. grant of the Flanders Research  Foundation (FWO). This research received funding from the Flemish Government under the ``Onderzoeksprogramma~Artificiële~Intelligentie~(AI)~Vlaanderen'' program.

\bibliographystyle{IEEEtran}
\bibliography{rfr}
\balance

\end{document}